\documentclass[10pt,journal,compsoc]{IEEEtran}
\usepackage{cite}
\ifCLASSINFOpdf
   \usepackage[pdftex]{graphicx}
\else
\fi
\usepackage{amsmath}
\usepackage{amssymb}
\usepackage{tabulary,xfrac,enumitem}
\usepackage[dvipsnames]{xcolor}
\usepackage[ruled,noresetcount]{algorithm2e}
\usepackage[pagebackref=false,breaklinks=true,colorlinks,bookmarks=false]{hyperref}
\definecolor{citecolor}{RGB}{34,139,34}

\usepackage{array}
\usepackage{url}

\usepackage{booktabs}
\usepackage{multirow}
\usepackage{threeparttable}
\usepackage{color, colortbl}
\usepackage{changes}
\usepackage{subcaption}
\usepackage{wrapfig}
\usepackage{diagbox}
\usepackage{arydshln} 
\usepackage{makecell}
\usepackage{etoolbox}     
\usepackage{caption}      
\usepackage{dblfloatfix}  

\makeatletter
\newcommand{\thickhline}{%
    \noalign {\ifnum 0=`}\fi \hrule height 0.5pt
    \futurelet \reserved@a \@xhline
}
\makeatother
\definecolor{LightGray}{gray}{0.9}
\definecolor{tabhead}{gray}{0.90}        
\definecolor{tabours}{RGB}{222,235,247}  
\newcommand{\ie}{\textit{i.e.}}
\newcommand{\eg}{\textit{e.g.}}

\usepackage{pifont}
\newcommand{\xmark}{\textcolor{red}{\ding{55}}}%
\newcommand{\cmark}{\textcolor{green}{\ding{51}}}%

\usepackage{graphicx}     
\usepackage{etoolbox}     

\makeatletter
\apptocmd{\@maketitle}{%
  \centering
  \includegraphics[width=\linewidth]{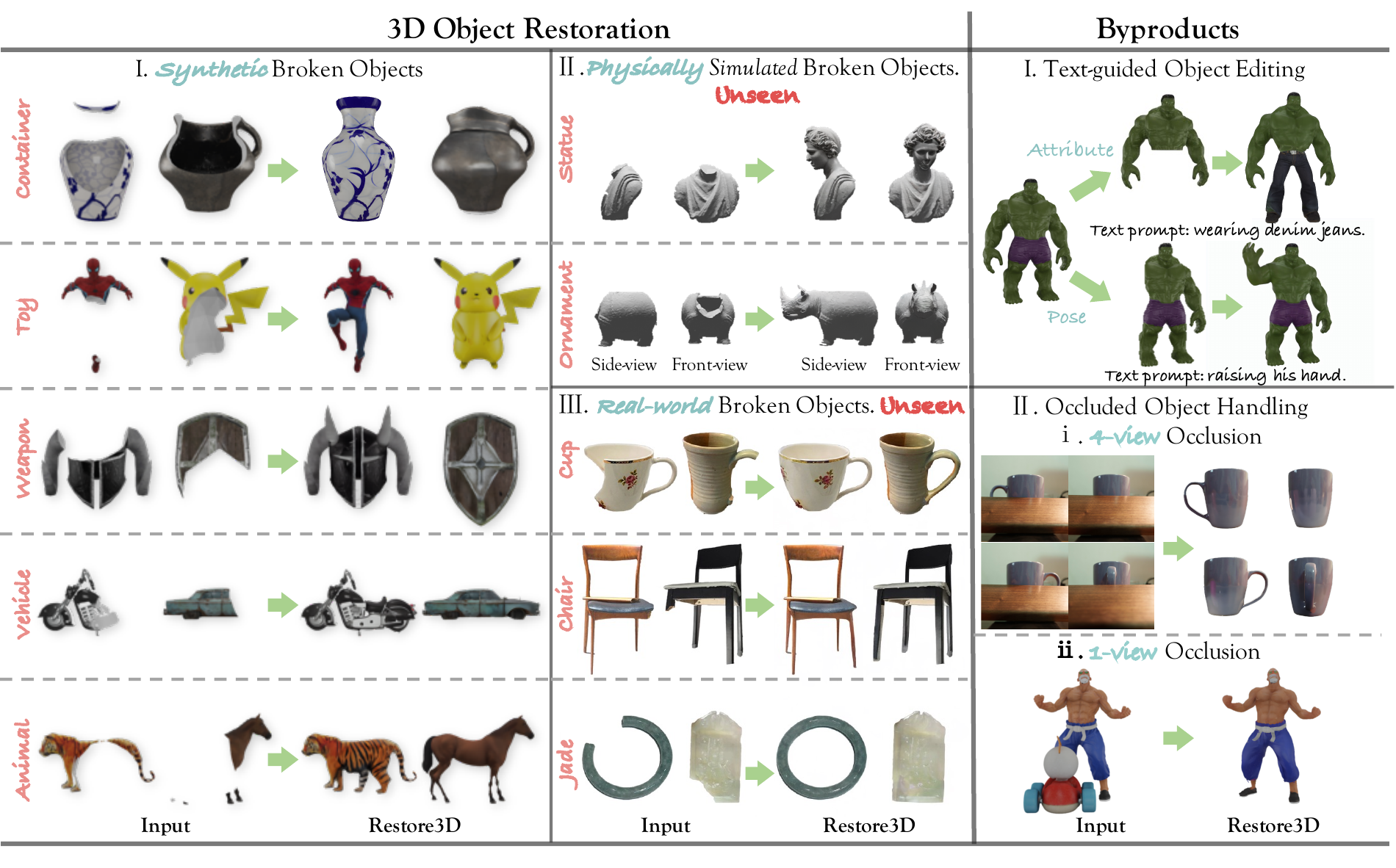}
  \par\vspace{-0.2em}
  \captionof{figure}{\textbf{Restore3D} targets preservation-aware shape-and-texture restoration of broken object-level 3D assets, producing plausible textured meshes for synthetic, simulated, and real broken objects.}
  \label{fig:first_f}
  \setcounter{figure}{0}
}{}{}
\makeatother

\begin{document}

%
\title{Restore3D: Breathing Life into Broken 
Objects with Shape and Texture Restoration}

%
%
%
%

\newcommand{\zongxin}[1]{{#1}}
\newcommand{\new}[1]{{#1}}

\author{Xiaolong~Shen,
        Zongxin~Yang,
        and~Yi~Yang,~\IEEEmembership{Fellow,~IEEE}
\thanks{
X. Shen, Z. Yang, and Y. Yang are with ReLER, CCAI, Zhejiang University, Hangzhou, China (Email: \{sxlongcs, yangzongxin, yangyics\}@zju.edu.cn). Y. Yang is the corresponding author.
}
}

%
%

\markboth{Restore3D}%
{Shen \MakeLowercase{\textit{et al.}}: Restore3D}

\IEEEtitleabstractindextext{%
\begin{abstract}
    Restoring incomplete or damaged 3D objects is crucial for cultural heritage preservation, occluded object reconstruction, and artistic design.
Existing methods primarily focus on geometric completion, often neglecting texture restoration and struggling with relatively complex and diverse objects.
We introduce Restore3D, a novel framework that simultaneously restores both the shape and texture of broken objects using multi-view images. To address limited training data, we develop an automated data generation pipeline that synthesizes paired incomplete-complete samples from large-scale 3D datasets. 
Central to Restore3D is a multi-view model, enhanced by a carefully designed Mask Self-Perceiver module with a Depth-Aware Mask Rectifier.
The rectified masks learned by the self-perceiver guide an image integration and enhancement phase, helping retain observed shape and texture patterns while refining the generated regions and mitigating the low-resolution limitations of the base model, yielding high-resolution, semantically coherent, and view-consistent multi-view images.
A coarse-to-fine reconstruction strategy is then employed to recover detailed textured 3D meshes from refined multi-view images. Experiments on synthetic and real broken-object benchmarks show that Restore3D improves multi-view restoration quality and textured-mesh reconstruction over representative inpainting, completion, and reconstruction baselines in the evaluated settings. Project Page: \url{restore3dx.github.io}
\end{abstract}

}


\maketitle
\setcounter{figure}{1}

\IEEEdisplaynontitleabstractindextext
%
\IEEEpeerreviewmaketitle

\IEEEraisesectionheading{\section{Introduction}\label{sec:introduction}}

\label{intro}
Recent advances in 3D generation and reconstruction techniques~\cite{cheng2023sdfusion,poole2022dreamfusion,lin2023magic3d, li2023instant3d, tang2024lgm} have demonstrated impressive capabilities, enabling a wide range of applications from virtual reality to digital manufacturing. Despite these strides, a fundamental yet underexplored problem persists: given a physically broken or incomplete 3D object, how can we recover the missing geometry and appearance while minimizing unintended changes to the intact portions? We term this problem \textbf{3D Object Restoration} --- the task of simultaneously reconstructing the complete shape and texture of a damaged 3D object, with the goal of retaining the geometry, color, and style of the originally observed regions as much as possible.

It is important to distinguish 3D Object Restoration from several superficially related tasks. \textit{Shape completion}~\cite{dai2017shape, rao2022patchcomplete} aims to fill in missing geometry from partial scans, yet it operates solely on geometry without considering texture. 
\textit{3D generation}~\cite{poole2022dreamfusion, lin2023magic3d} creates novel objects from reference images or text prompts, with no mechanism to anchor the output to an existing incomplete object. In contrast, 3D Object Restoration requires \textbf{joint shape-and-texture recovery} conditioned on a given broken object, with the defining constraint that the intact portions should be \textbf{preserved} while the missing or damaged regions are restored --- a requirement absent from all of the above tasks. This makes 3D Object Restoration an underexplored setting with a different combination of requirements from prior shape completion, 3D generation, and 3D inpainting tasks.


Key challenges in achieving this goal include: \textit{i) Data Scarcity}. Existing 3D datasets~\cite{chang2015shapenet, dai2017shape, rao2022patchcomplete} focus primarily on shape completion of simple objects, often neglecting texture. Creating diverse, paired broken-complete data remains labor-intensive and time-consuming. \textit{ii) Complexity of Restoration}. Real-world broken objects exhibit diverse fracture patterns and varying complexity. Simpler methods work only for limited categories of simple objects; when applied to more complex cases, the synthesized regions fail to align with the original parts, or worse, parts of the original structure are overwritten or discarded during restoration. \textit{iii) Preservation-Aware Restoration}. Unlike generation or completion tasks that produce entirely new outputs, restoration should avoid unnecessary changes to intact portions, including their color, style, geometric details, and structural coherence. Unintended drift in observed regions can reduce restoration fidelity, especially near the boundary between preserved and generated regions.


To address these challenges, we propose several complementary solutions: \textbf{i) Synthetic Data Generation}. To overcome the limitations of existing datasets, we propose to synthesize paired broken and complete data. \textbf{ii) Leveraging Foundation Models}. Recent advancements in foundation models~\cite{hong2023lrm, shi2023mvdream, rombach2022highresolution, oquab2023dinov2, kirillov2023segany, depth_anything_v2} have demonstrated exceptional generalizability, due to their extensive architectures, large-scale datasets, and adaptability through fine-tuning. We incorporate foundation models to provide prior knowledge, enabling our framework to effectively handle complex and diverse cases. \textbf{iii) Task-Specific Structures}. While foundation models offer valuable priors, task-specific components are necessary to tailor their application. Motivated by studies~\cite{zhang2023adding, ye2023ip-adapter, mou2023t2i}, we guide these models toward optimal probability distributions with specialized modules, achieving more accurate and contextually appropriate restorations.

 \begin{figure*}[!t]
		\centering
		\includegraphics[width=0.95\textwidth]{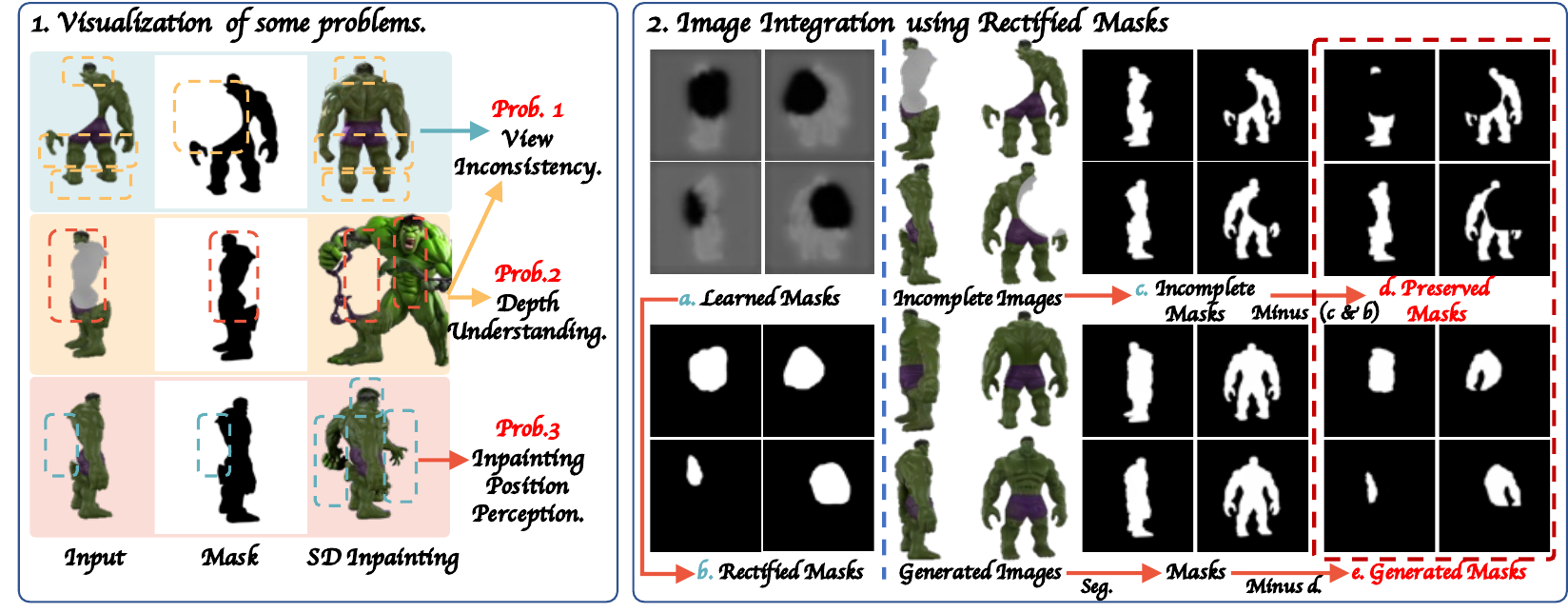}

		\caption{
            \textbf{The importance of masks.} 
            In single-view inpainting, user-provided masks define the regions requiring inpainting. However, in a multi-view context, manually creating consistent masks across all views is impractical. Directly inverting object masks to serve as inpainting masks inevitably causes issues (see Prob. 1 \& 3). Moreover, manually adjusting masks based on depth information (see Prob. 2) is labor-intensive and time-consuming. As shown in the right figure (a), our mask self-perceiver can automatically indicate the regions that need to be completed.   
            By leveraging both preserved and generated masks (d \& e), our approach retains the object's patterns, ensuring accurate and consistent multi-view inpainting. These masks are also used for the image enhancement stage to yield high-resolution restored images (see Fig.~\ref{enhance_pipe}).
		}
  
    \label{mi_mo}
\end{figure*}

Concretely, we first produce an automatic pipeline to construct paired data, which uses the Boolean modifier in Blender. It offers diverse and large-scale data that are difficult to acquire manually.
Second, we propose an innovative framework named \textbf{Restore3D}, comprising two key components, \ie, \textbf{multi-view image inpainting and reconstruction}. There are several foundational models~\cite{shi2023mvdream, liu2023zero,xu2024instantmesh} in these two components that we can leverage prior knowledge to further handle more diverse incomplete objects effectively. However, simply applying foundational models to multi-view images introduces several \textbf{challenges}, as shown in Fig.~\ref{mi_mo}, including:
\textit{1) View Inconsistency}: Generated results often differ across views, leading to visual incoherence.
\textit{2) Depth Understanding}: Existing models often lack robust depth perception, resulting in failures to recognize occlusions and spatial relationships.
\textit{3) Inpainting Position Perception}: Accurately identifying regions requiring inpainting can be difficult, especially for large masks.

To address these issues, we propose a \textbf{multi-view} base model combined with a specially designed \textbf{mask self-perceiver} module incorporating a \textbf{depth-aware mask rectifier}. This module autonomously perceives and reconstructs missing components, preserving the integrity of original broken regions and ensuring consistent results across multiple views. Additionally, by leveraging the preserved and generated masks predicted by the self-perceiver, we can develop an image integration and enhancement pipeline (see Fig.~\ref{mi_mo} \& \ref{enhance_pipe}), yielding high-quality and consistent results. To convert high-quality multi-view images into 3D objects, we employ large reconstruction models (LRMs)\cite{hong2023lrm,xu2024instantmesh}, which offer efficient single- and multi-view object reconstruction capabilities. To overcome the limitation of coarse outputs from these models, we adopt a coarse-to-fine refinement approach. Leveraging recent advances in surface normal prediction models\cite{bae2024dsine, ye2024stablenormal}, we inject normal priors to progressively enhance geometric quality, and refine texture based on updated geometry by using enhanced images. This ensures that our refined shapes and textures maintain high fidelity, even for complex scenarios.

We conduct extensive experiments on Objaverse \cite{deitke2023objaverse}, GSO \cite{downs2022googlescannedobjectshighquality},
Breaking Bad Dataset~\cite{BBD_dataset},
Fantastic Breaks~\cite{Lamb_2023_CVPR} and OmniObject3D \cite{wu2023omniobject3d} to validate the quality of inpainting and reconstruction. The results show that Restore3D achieves stronger inpainting quality than representative 2D and multi-view inpainting baselines~\cite{lugmayr2022repaintinpaintingusingdenoising, zhang2023adding, rombach2022highresolution, weber2023nerfiller} in our restoration setting. Notably, Instant3dit~\cite{barda2024instant3ditmultiviewinpaintingfast} improves substantially when using our predicted masks, suggesting that restoration-region perception is a key bottleneck for this task.
By carefully designing a mask self-perceiver, our method can alleviate view inconsistency, understand depth concepts, and capture inpainting regions, achieving consistent structure and texture styles without requiring user-provided masks to indicate inpainting regions. 
For reconstruction, our approach enhances both geometric and texture quality as shown in Fig.~\ref{fig:first_f}, indicating that our proposed framework is capable of producing complete shapes and textures with relatively high fidelity compared to baseline methods~\cite{openlrm,xu2024instantmesh, xiang2024structured}. Overall, our contributions are summarized as follows,
\begin{itemize}
\item We study preservation-aware 3D object restoration, which jointly recovers missing geometry and texture for broken object-level 3D assets while retaining observed regions. To the best of our knowledge, Restore3D is the first framework that combines automatic restoration-region perception, multi-view shape-and-texture restoration, and textured mesh reconstruction for this setting. To support this task, we introduce an automated data synthesis pipeline that generates large-scale paired broken-complete data, yielding a dataset named RestoreIt-3D.
\item We propose Restore3D, a mask-aware restoration framework that decomposes broken-object restoration into restoration-region perception, multi-view image restoration, image integration/enhancement, and coarse-to-fine textured mesh reconstruction. We design a mask self-perceiver with a depth-aware mask rectifier to autonomously identify restoration regions and reduce unintended changes to the original observed features. An image integration and enhancement pipeline further restores fine details, and a coarse-to-fine reconstruction strategy with normal priors ensures high-fidelity geometry and texture.
\item Experiments on multiple synthetic, simulated, and real broken-object benchmarks show that Restore3D improves restoration and reconstruction quality over representative baselines, with ablations supporting the role of mask perception, image integration, and coarse-to-fine refinement.
\end{itemize}
\section{Related Work}

\noindent\textbf{2D Inpainting and Generation models.} 2D inpainting methods are designed to complete missing content in an image using a given image and mask. LaMa \cite{suvorov2021resolution} utilizes fast Fourier convolutions, a large receptive field, and extensive training masks to effectively fill large missing areas, producing plausible inpainting results. 
Recent advancements in image generation \cite{rombach2022highresolution, zhang2023adding} have demonstrated superior performance and can be adapted for inpainting tasks with high-quality outcomes. RePaint~\cite{lugmayr2022repaintinpaintingusingdenoising} modifies the diffusion generation process, allowing it to be used for inpainting. 
NeRFiller~\cite{weber2023nerfiller} uses grid priors to make the 2D diffusion model produce more consistent multi-view inpainting results. Instant3Dit~\cite{barda2024instant3ditmultiviewinpaintingfast} employs a multi-view inpainting model combined with a large reconstruction model to enable rapid editing of 3D objects. Many of these methods rely on user-specified or externally provided masks to indicate the regions to be modified, whereas restoration of broken objects requires inferring restoration regions from the damaged input itself.

\noindent\textbf{3D Inpainting and Multi-view Enhancement.}
Closely related to our task are 3D inpainting methods, including NeRF-based scene inpainting methods such as SPIn-NeRF~\cite{spinnerf} and MVIP-NeRF~\cite{mvipnerf}, and recent 3D Gaussian inpainting methods such as 3DGIC~\cite{gaussian3dgic}. These methods aim to complete masked regions in scene-level radiance-field or Gaussian-splatting representations, often assuming user-provided masks and focusing on object removal or scene completion. Recent multi-view enhancement and reconstruction methods, such as MVD$^2$~\cite{mvd2}, 3DEnhancer~\cite{luo20253denhancerconsistentmultiviewdiffusion}, and Sharp-It~\cite{sharpit}, improve multi-view consistency and high-frequency details for 3D generation. In contrast, Restore3D focuses on broken object-level restoration, where the method must infer restoration regions automatically, retain observed object parts, recover missing geometry and texture, and output a textured mesh.

 \begin{figure*}[!t]
		\centering
		\includegraphics[width=1.0\textwidth]{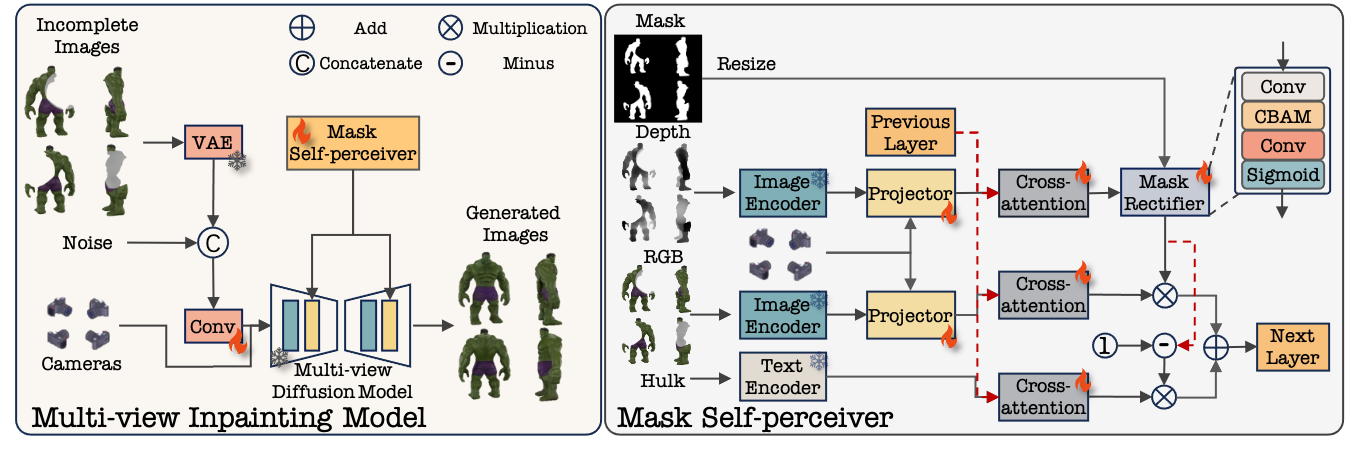}
        
		\caption{
            \textbf{Multi-view Image Inpainting.} 
            We carefully design a mask self-perceiver based on a multi-view diffusion model that composes the image and text features with a spatial mask predicted by a depth-aware mask rectifier, therefore the model can automatically perceive the missing part and further generate it meanwhile preserving the original parts.
		}
  
		\label{pipe_multi_inpa}
  
\end{figure*}

\noindent\textbf{3D Generation and Completion.}
Recent 3D generation models~\cite{wang2023prolificdreamer,lin2023magic3d, chen2023fantasia3d} showcase promising results. DreamFusion~\cite{poole2022dreamfusion} and SJC~\cite{wang2023score} are first proposed to generate 3D assets from text using the strong 2D text-to-image generation model~\cite{rombach2022highresolution}. 
As 2D diffusion models easily lead to 3D inconsistency,  some works~\cite{liu2023zero, zhou2023sparsefusion,tang2023mvdiffusion,szymanowicz2023viewset,tewari2023diffusion,xu2023dmv3d} focus on consistent multi-view image diffusion models.
MVDream~\cite{shi2023mvdream} uses 3D self-attention and camera embedding to achieve multi-view text-to-image generation.
Considering the time-consuming nature of SDS-based methods, there are some works~\cite{hugging2023one2345,long2022sparseneus,li2023instant3d,long2023wonder3d,tang2024lgm,wu2024unique3d, lu2024direct2} that use multi-view diffusion models and reconstruction models. Another line for 3D generation is that directly train 3D generative models using 3D representations like point cloud~\cite{nichol2022point, zeng2022lion, luo2021diffusion}, meshes~\cite{liu2023meshdiffusion,gao2022get3d}, neural fields~\cite{ kim2023neuralfield, anciukevivcius2023renderdiffusion, muller2023diffrf, jun2023shape, zhang20233dshape2vecset, erkocc2023hyperdiffusion, chen2023single}.
In addition to 3D generation, recent 3D shape completion works~\cite{kasten2023pointcloud,zhang2021unsupervised,dai2019scan2mesh,autosdf2022,pan2021variational, cheng2023sdfusion, chu2023diffcomplete} usually use different types of 3D representations and networks to model global and local structures, \eg, point cloud, sdf, GAN, VAE, and diffusion models. Most shape completion methods focus primarily on geometry recovery from partial 3D observations. They are therefore complementary to our setting, which requires joint geometry-and-texture restoration of damaged objects and preservation of observed regions.

\noindent\textbf{Texture Generation.}
Several texture generation works~\cite{richardson2023texture,cao2023texfusion,chen2023text2tex} use an iteratively texturing strategy based on the pre-trained depth-to-image diffusion models, yielding high-quality texture.
However, these methods tend to error lighting inherited from training data.
Paint3D~\cite{zeng2023paint3d} proposes a shape-aware UV Inpainting and a shape-aware UVHD diffusion model to alleviate this situation.
There is another line to learn texture.
Texturify~\cite{siddiqui2022texturify} employs texture maps on the surface of meshes and uses StyleGAN~\cite{karras2019style} to predict texture. Mesh2Tex~\cite{bokhovkin2023mesh2tex} incorporates an implicit texture field for texture prediction. These methods are lacking in global information modeling. PointUV~\cite{yu2023texture} first trains a diffusion model specifically for mesh texture generation, and the proposed coarse-to-fine framework allows it to enjoy the efficiency of 2D representation while enhancing 3D consistency. Other approaches like AUV-net~\cite{chen2022auv}, LTG~\cite{yu2021learning}, and TUVF~\cite{cheng2023tuvf} learn to generate UV-Maps for 3D shapes. However, they typically focus on the texture generation starting from a complete shape.

\section{Methodology}
\subsection{Preliminary}

\noindent\textbf{Multi-view Diffusion models.} 
Extending 2D generation models to the multi-view domain has been explored in various works~\cite{liu2023zero,shi2023mvdream}. These extensions often incorporate modifications like adding camera conditions and adjusting the attention mechanisms to enable effective multi-view synthesis. 
In this paper, we adopt MVDream as our base model. MVDream modifies the spatial attention mechanism in Stable Diffusion~\cite{rombach2022highresolution}, allowing the attention to focus on corresponding features across different views. 

\subsection{Data Preparation \& Task Definition}
\textbf{Motivation.} We browse the datasets of related tasks and find that the existing datasets~\cite{chang2015shapenet, deitke2023objaverse,collins2022abo} are not sufficient to handle the shape and texture completion of broken objects, which suggests the need to construct specific broken and complete paired data. However, collecting large-scale paired data in the real world is \textit{time-consuming and labor-intensive}. Thus we propose to \textit{synthesize} broken and complete paired data. 

\noindent\textbf{Data Collection.} We select the recent dataset, G-objaverse~\cite{qiu2023richdreamer} that has \textit{more diverse and general objects}, and sample about 83K 3D objects from this dataset.

\noindent\textbf{Synthesis Pipeiline.}
Specifically, we propose an automatic data processing technique using Boolean operations (\ie, Difference and Intersect) of Blender. 
Additionally, we equip the dataset with text captions using Cap3D~\cite{luo2023scalable3dcaptioningpretrained}.
Subsequently, we normalize and merge the prepared 3D data.
The use of Boolean operations requires the introduction of another object. Therefore, we use an ico sphere or cube with random size and rotation angle and then randomly place them inside the 3D bounding box of the prepared 3D data to ensure that the objects can be realistically segmented. After that, it is essential to render this processed data in the format of RGB images to facilitate model learning. We execute the rendering at a resolution of 256$\times$256. 
The camera settings include a randomly chosen elevation between -10$^{\circ}$ and 30$^{\circ}$. Additionally, the azimuth values are uniformly rendered from 0$^{\circ}$ to 360$^{\circ}$ with a randomly sampled start view, producing a total of 32 images per object. The Fov of the camera is randomly from 35$^{\circ}$ to 45$^{\circ}$ and the distance is always 2.

\noindent\textbf{Task Definition.} 
The 3D object restoration task aims to reconstruct a complete 3D mesh with texture from multi-view images of a damaged object.
Given \textbf{multi-view images} \( \{I_1, I_2, \dots, I_n\} \) capturing a damaged object from different angles and corresponding \textbf{camera parameters} (\( K_i \), \( E_i \)) , the model will output a complete 3D mesh \( M = (V, F, T) \): Vertices, Faces, and Textures.

\subsection{Multi-view Image Inpainting}
\noindent\textbf{Motivation.} Traditional single-view image inpainting methods~\cite{suvorov2021resolution, rombach2022highresolution, zhang2023adding} rely on the user-provided masks that indicate the areas to be inpainted. 
While this approach works well in the context of single-view images, it presents significant challenges when extended to multi-view contexts as shown in Fig.~\ref{mi_mo}. 
\textit{1. View inconsistency.} In a multi-view scenario, the user is required to manually provide a mask for each of the views (\eg, four views in our case). This also introduces the risk of errors, as the mask needs to be accurately aligned across different perspectives to maintain 3D consistency.
\textit{2. Uncertainty Regarding Inpainting Areas.} These models cannot autonomously perceive the regions that require inpainting when a large mask is applied. Additionally, they do not incorporate depth perception, limiting their understanding of occlusion and spatial relationships. 
To address these challenges, we propose an innovative approach that enables the model to \textit{ensure view consistency} and \textit{self-perceive the mask}. Concretely, we design the following two parts.

\noindent\textbf{Mask Self-perceiver.} We propose a mask self-perceiver module based on a multi-view image generation model as shown in Fig.~\ref{pipe_multi_inpa}. It has two projectors that consist of transformer-based blocks and camera modulation layers, which project the depth and image features~($f_d, f_{r}$) extracted from CLIP~\cite{radford2021learningtransferablevisualmodels} to the diffusion feature space. The camera modulation helps the model to discriminate the feature under different cameras. Then these projected features~($p_d, p_r$) will be fed to the respective cross-attention blocks as key and value~($\mathbf{K_d, K_r, V_d, V_r}$). The process can be formulated as follows, where $f_*$ can be depth or image features, $p_*$ is the projected features of them.
\begin{equation}
    p_{*} = \mathbf{Proj}(f_*, c) = \mathbf{Trans}(\mathbf{Mod}(f_*, c))
  \label{eq3}
\end{equation}
\begin{equation}
    s_* = \mathbf{Softmax}(\frac{\mathbf{QK_*^T}}{\sqrt{d}})\mathbf{V_*}
  \label{eq2}
\end{equation}
 Similarly, $s_*, \mathbf{K_*}$ and $\mathbf{V_*}$ are the results of $p_*$ via cross-attention and linear layers. $\mathbf{Q}$ originates from the pre-layer features in the diffusion model.

\noindent\textbf{Depth-aware Mask Rectifier.} Since depth effectively captures the incomplete shape while disregarding texture information, the rectifier can focus solely on identifying the regions that require generation and preservation. Moreover, the depth can help the model understand the spatial relation and occlusion. Specifically, This module leverages depth features obtained after the cross-attention layer, along with incomplete masks, and inputs them into a mask rectifier.  The rectifier then outputs a mask indicating where needs to be generated \ie, leveraging the text features and where needs to be preserved \ie, using the image features. The process can be formulated as follows, 
\begin{equation}
    \small
    \mathcal{M}_{r} = \mathbf{Sigmoid}(\mathbf{Conv}(\mathbf{CBAM}(\mathbf{Conv}[s_d, \mathcal{M}_o])))
  \label{eq1}
\end{equation}

\begin{equation}
    f_n =  (\mathbf{1} - \mathcal{M}_r) s_t+ \mathcal{M}_r s_r 
  \label{eq4}
\end{equation}
$\mathbf{Conv}$ is convolution layers, $\mathbf{CBAM}$ is Convolutional Block Attention Module~\cite{woo2018cbamconvolutionalblockattention}.

\noindent\textbf{Training objectives} 
Given training samples, including incomplete images~$\mathcal{I}$, depth images $\mathcal{D}$, incomplete masks $\mathcal{M}$, text prompts $\mathcal{P}$ and camera embedding $\mathcal{C}$, the multi-view inpainting loss can be formulated as follows, 
\begin{equation}
    \small
     \mathcal{L} = \min_{\theta}\mathbb{E}_{z, \epsilon \sim \mathcal{N}(\mathbf{0,I}), t}\Vert \epsilon -\epsilon_\theta(z_t; t, \mathcal{I}, \mathcal{D}, \mathcal{M}, \mathcal{P}, \mathcal{C}) \Vert^2_2.   
\end{equation}


\begin{figure}[t]
  \centering

  \includegraphics[width=\linewidth]{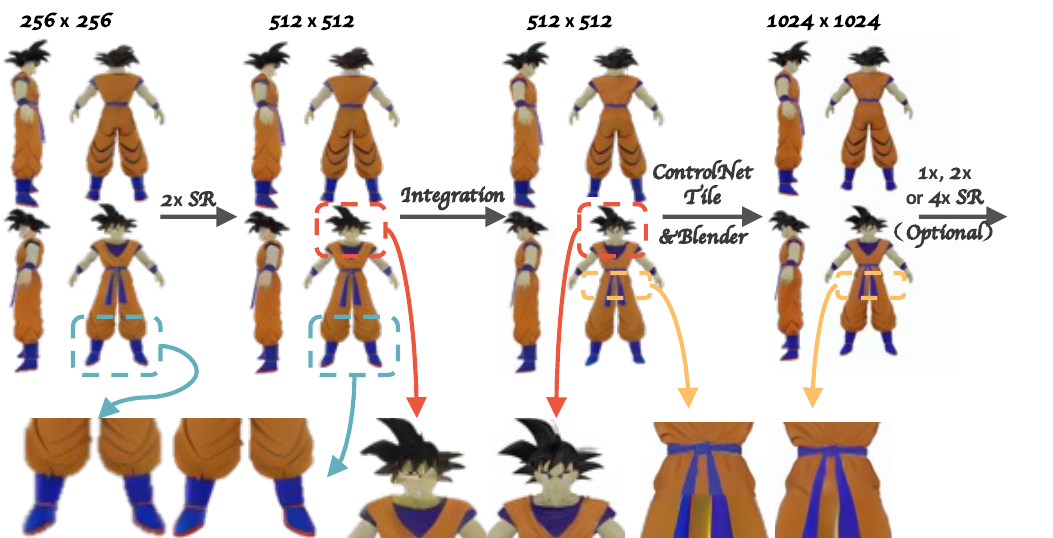}
      
  \caption{
          \textbf{Image Integration and Enhancement Pipeline using Rectified Masks.}
  }
  \label{enhance_pipe}

\end{figure}

\subsection{Image Integration and Enhancement}
\noindent\textbf{Motivation. } The input resolution of multi-view model is 256 x 256, which is subsequently encoded to 32 x 32 using a Variational Autoencoder. As a result, \textit{local details are compressed, leading to a loss of clarity in both the original and generated regions of the image.} This compression often causes the inpainted part to be unclear, and the reconstructed image may lose fine details that are essential for achieving high-quality results. Moreover, \textit{high-quality images} will \textit{help} the next \textit{reconstruction} stage to give accurate and detailed textured meshes. To address these challenges, we propose a pipeline that enables the model to \textit{restore local details and preserve the original patterns}. 
 \begin{figure*}[!t]
		\centering
		\includegraphics[width=1.0\textwidth]{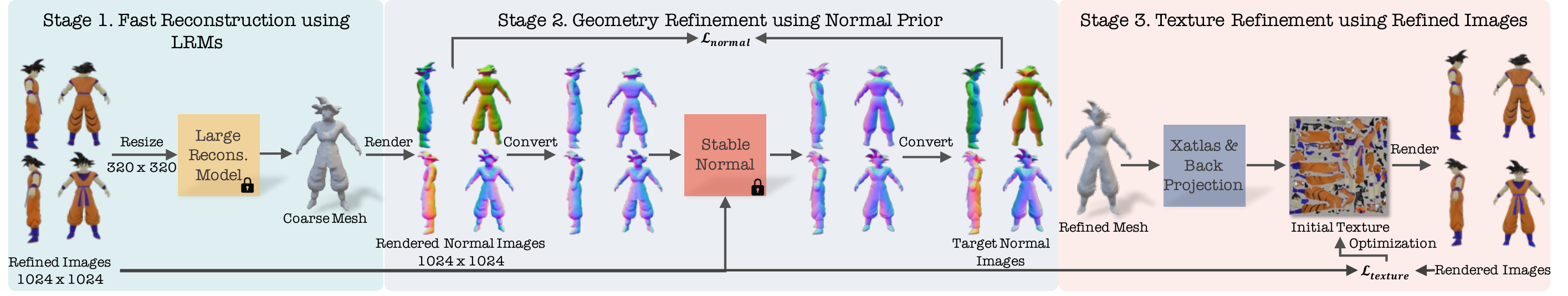}
        
		\caption{
            \textbf{Geometry and Texture Refinement.} We separately refine the geometry and texture of the coarse results inferred by LRMs~\cite{xu2024instantmesh}.
		}
        
 
		\label{mesh_refine}
  
\end{figure*}

\noindent\textbf{Enhancement Models.} 
We explore two types of enhancement models.
\textit{Real-ESRGAN}~\cite{wang2021realesrgan} is effective at preserving the patterns of low-resolution images with minimal misalignment, making it ideal for recovering the overall structure.
\textit{ControlNet-Tile}~\cite{zhang2023adding} offers advanced capabilities for enhancing image details, but will modify the original pattern when a high denoising step is used.
Based on these properties, we design the following enhancement pipeline. \textit{1. Input resolution alignment using Real-ESRGAN.} Before integrating with the original images, we need to align the resolution. Using Real-ESRGAN effectively preserves the overall structure and does not introduce content that is not related to the original style. 
\textit{2. Integration of generated and original parts using rectified masks.}
As depicted in Fig.~\ref{enhance_pipe}, this procedure infers the preserved and generated masks used to compose the images, which preserves the original parts as soon as possible. However, this procedure inevitably leads to some artifacts, \eg,  inconsistent color transitions. To address these artifacts, we leverage the mentioned property of ControlNet-Tile to enhance the images.
\textit{3. Image harmonizing using ControlNet-Tile with a blending strategy.} 
Directly using ControlNet-Tile will alter the original pattern and destroy the integration step. Inspired by previous works \cite{Avrahami_2022_CVPR, lugmayr2022repaintinpaintingusingdenoising}, we incorporate a mask blending technique within the diffusion process. This technique helps maintain the original patterns, eliminates any gaps caused by integration in image space, and enhances the image quality. 


\subsection{Multi-view Image Reconstruction}
\noindent\textbf{Fast Reconstruction using Large Reconstruction Models (LRMs).} 
Recent advancements in LRMs~\cite{hong2023lrm, tang2024lgm, xu2024instantmesh}, which leverage sophisticated architectures, large-scale datasets, and extensive model parameters, have demonstrated impressive capabilities in 3D object reconstruction from single or sparse-view images. 
These models are particularly well-suited for tasks requiring fast mesh reconstruction. However, while LRMs can produce initial reconstructions efficiently, the results are often \textit{coarse and lack the fine details} necessary for high-quality 3D representations. To address this limitation, we adopt a coarse-to-fine schema and refine the shapes and textures of the outputs generated by LRMs, separately, as shown in Fig.~\ref{mesh_refine}. 

\noindent\textbf{Geometry Refinement using Normal Prior.} A key component in optimizing shape structure is to obtain high-quality surface normals. Recent surface normal estimation methods~\cite{ye2024stablenormal} have demonstrated the ability to predict relatively accurate normals for in-the-wild monocular images or videos. Therefore, we can employ an \textit{off-the-shelf} normal estimation model to provide normal priors and then use it to optimize the shape structure of 3D objects.
Since these models are primarily trained on monocular images or videos, the predicted normals are typically in camera space. Thus we need to convert these normals into world space using camera extrinsic parameters. Specifically, we select StableNorm, a model that accepts coarse rendered normals and RGB images as inputs to predict refined normal outputs. The consistency of the rendered normals contributes to the stability and accuracy of the predicted normals, allowing for more precise geometry refinement.



\noindent\textbf{Texture Refinement using High-quality Images.} 
Since the current shape differs from the coarse shape, the original texture no longer aligns with the updated geometry. Thus we propose to learn the textures that better match the optimized shape.
Concretely, we can use Xatlas to obtain UV coordinates, enabling us to back-project the colors from the inpainted images onto the UV textures. After that, we treat the UV textures as parameters and use the high-quality images to optimize it.


\noindent\textbf{Training Objectives.} We apply a normal loss~$\mathcal{L}_{normal}$ based on the rendered normals $\mathcal{I}_n$ and the target normals $\hat{\mathcal{I}_n}$. Additionally, we apply a mask loss~$\mathcal{L}_{mask}$ to ensure that the optimization regions are correctly aligned. The loss function is defined as follows,
\begin{equation}
    \mathcal{L}_{shape} = \mathcal{L}_{normal} + \mathcal{L}_{mask}
    = \Vert \mathcal{I}_{n}  -  \hat{\mathcal{I}_{n}}  \Vert^2_2 + \Vert \mathcal{M} - \hat{\mathcal{M}} \Vert^2_2    .
\end{equation}
To optimize the texture, we use a RGB loss~$\mathcal{L}_{rgb}$ on the rendered images $\mathcal{I}_{rgb}$ and enhanced images $\hat{\mathcal{I}_{rgb}}$. The mask loss $\mathcal{L}_{mask}$ is also applied. Moreover, the SSIM~$\mathcal{L}_{ssim} $ loss is introduced to improve the texture quality. The loss functions are defined as follows, 
\begin{equation}
    \begin{aligned}
          \mathcal{L}_{tex} &= \mathcal{L}_{rgb} + \mathcal{L}_{mask} + \lambda\mathcal{L}_{ssim} 
    \\ &= \Vert \mathcal{I}_{rgb}  - \hat{\mathcal{I}_{rgb}}   \Vert^2_2  + \Vert \mathcal{M} - \hat{\mathcal{M}} \Vert^2_2    + \lambda\mathbf{SSIM(\mathcal{I}, \hat{\mathcal{I}})},
    \end{aligned}
\end{equation}
where $\lambda$ is a weight parameter.

 \begin{figure*}[!t]
		\centering
		\includegraphics[width=0.99\textwidth]{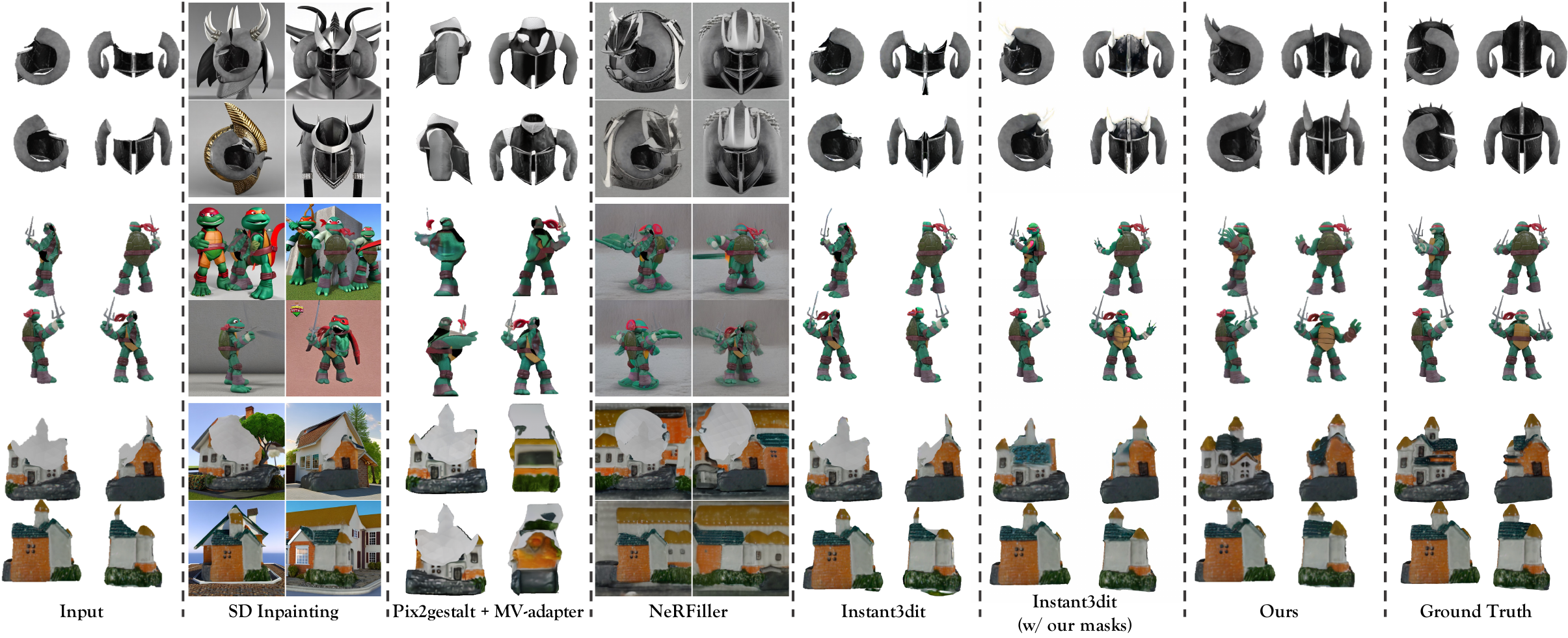}
        
		\caption{
            \textbf{Visual Comparison with Inpainting Methods.}}
		
		\label{vs_inpainting}
        
\end{figure*}

\section{Implementation Details}
\label{imp}
We train the multi-view inpainting model using four NVIDIA A100 GPUs. We use the Adam optimizer and incorporate classifier-free guidance. The training is conducted with a learning rate of 1e-4 and a batch size of 256. MVDream is utilized as the base model for multi-view inpainting, while InstantMesh is employed as the large reconstruction model. 
The input consists of 4-view images. For the sampling process, we employ DDIM with 50 steps and a guidance scale of 5.0.

\section{Experiments}
\label{exppp}

\noindent\textbf{Dataset.} For model training, we sample approximately 83K data from the G-objaverse dataset ~\cite{qiu2023richdreamer} and process them using our proposed pipeline. For model testing, we sample approximately 350 data from the GSO~\cite{downs2022googlescannedobjectshighquality}, Omniobject~\cite{wu2023omniobject3d}, and Objaverse~\cite{deitke2023objaverse} datasets. 
We also test our model on the Breaking Bad Dataset~\cite{BBD_dataset} and Fantastic Breaks~\cite{Lamb_2023_CVPR}, which include physically simulated and real-world broken objects, to evaluate its generalizability. 

\noindent\textbf{Metrics.} To assess image quality, we choose Peak Signal-to-Noise Ratio (PSNR), Frechet Inception Distance (FID), Learned Perceptual Image Patch Similarity (LPIPS), and Structural Similarity Index Measure (SSIM). We evaluate geometry quality using Chamfer Distance (CD) and F-scores. These metrics evaluate overall restoration and reconstruction quality. Since preservation of observed regions is a central goal of restoration, our ablations on mask prediction and image integration further analyze how the proposed components reduce unintended changes to preserved regions.


 \begin{figure*}[!t]
\centering
  
		\includegraphics[width=0.99\textwidth]{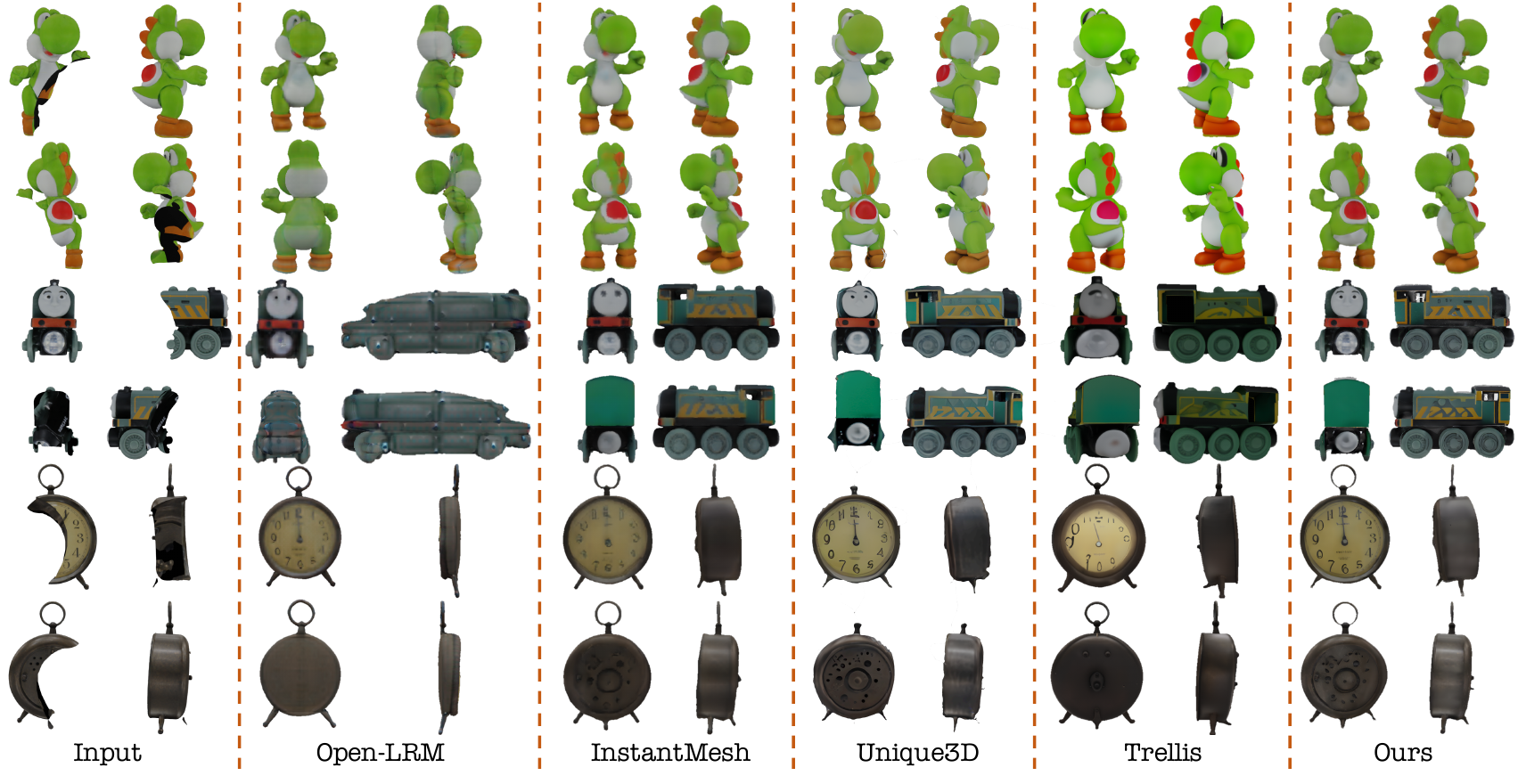}
        
		\caption{
            \textbf{Visual Comparison with Reconstruction Models.}
		}
		\label{rec_ql}
        
\end{figure*}


\begin{table}
    \centering
    \caption{\textbf{Comparison with Inpainting Methods.} $\triangle$ means using Depth-Anything~\cite{depth_anything_v2} to obtain the depth images. $\clubsuit$ means using MV-adapter\cite{huang2024mvadaptermultiviewconsistentimage}. $\heartsuit$ means using our model's predicted masks as inpainting masks.}
    \begin{tabular}{l | c  c c c}
\toprule[2pt]
\rowcolor{tabhead}
Method & PSNR $\uparrow$ & LPIPS  $\downarrow$ & FID  $\downarrow$ & SSIM  $\uparrow$\\
\midrule[1pt]
Repaint & 10.55 & 0.31&69.57 & 0.76\\
SD  & 12.58 & 0.22& 61.15 & 0.83  \\
ControlNet  & 10.66 & 0.30&69.91 & 0.76\\
Pix2gestalt $\clubsuit$ &16.43	&0.21 &75.08 &0.86 \\
MVInpainter	&11.12	&0.29 & - &	0.79 \\
NeRFiller & 12.03 & 0.25 & 65.20 & 0.82 \\
Instant3dit &	19.40	&0.10	&48.03 &0.94 \\
Instant3dit $\heartsuit$ &22.37	&0.07 &36.08 &0.95 \\
Ours $\triangle$ & 25.29 & 0.07 & 32.05 & 0.95\\
\rowcolor{tabours}
Ours & \textbf{25.50} & \textbf{0.06} & \textbf{31.82} & \textbf{0.95}\\
\bottomrule[1pt]
\end{tabular}
\label{inpainting_res}
\end{table}

\begin{table}
\centering
\caption{\textbf{Comparison with Reconstruction Methods.} }
\begin{tabular}{l | c  c c c}
\toprule[2pt]
\rowcolor{tabhead}
Method & PSNR $\uparrow$ & LPIPS  $\downarrow$ & CD  $\downarrow$ & F-Score  $\uparrow$\\
\midrule[1pt]
Open-LRM & 16.90 &	0.15 &0.011 & 0.179 \\
InstantMesh & 20.60	&0.11 & 0.006 & 0.321\\
Unique3D		&22.00	&0.14 &0.005	&0.306 \\
Direct3D		&-	&- &0.006	&0.297 \\
Trellis		&21.78	&0.12 &0.005	&0.335 \\ 
Hunyuan3D-2	&21.31	&0.14	&0.006	&0.346 \\
Amodal3R	&19.37	&0.15	&0.008	&0.248\\
\rowcolor{tabours}
Ours & \textbf{23.35}	& \textbf{0.09} & \textbf{0.005} & \textbf{0.389} \\
\bottomrule[1pt]
\end{tabular}
\label{rec_res}
\end{table}

\subsection{Inpainting Results.}

\noindent\textbf{Baselines.} We compare our method with single-view image inpainting, \ie, Repaint\cite{lugmayr2022repaintinpaintingusingdenoising}, Stable-Diffusion~\cite{rombach2022highresolution}, Controlnet~\cite{zhang2023adding}, \ie, Pix2gestalt + MV-adater~\cite{ozguroglu2024pix2gestaltamodalsegmentationsynthesizing, huang2024mvadaptermultiviewconsistentimage}
and multi-view inpainting methods, \ie, Nerfiller~\cite{weber2023nerfiller},  MVInpainter~\cite{cao2024mvinpainterlearningmultiviewconsistent} and Instant3dit~\cite{barda2024instant3ditmultiviewinpaintingfast}. Note that we do not use the image integration and enhancement pipeline for a fair evaluation.

        
  

\noindent\textbf{Qualitative Comparison.} 
As shown in Fig.~\ref{vs_inpainting}, the results demonstrate that our model produces plausible and coherent inpainting outcomes.  
Previous methods require user-provided masks to guide the model in generating missing parts. 
When given a relatively large mask, these methods struggle to capture the inherent structure of the objects, leading to less accurate and coherent inpainting. 
In contrast, our approach does not require predefined inpainting masks. It autonomously perceives and reconstructs missing regions,
capturing the underlying structure of the object.
This capability allows our method to produce high-quality and structurally consistent inpainting results.

\begin{table*}[t]
\centering
\begin{minipage}[t]{0.48\textwidth}
\centering
\small
\setlength{\tabcolsep}{6pt}
\caption{\textbf{Generalization Ability on Real-world Dataset~\cite{Lamb_2023_CVPR}.}}
\begin{tabular}{l | c c c}
\toprule[2pt]
\rowcolor{tabhead}
Method & PSNR $\uparrow$ & LPIPS  $\downarrow$ & SSIM  $\uparrow$\\
\midrule[1pt]
SD & 12.59 & 0.72 & 0.40  \\
Controlnet & 15.63 & 0.55 & 0.56 \\
Nerfiller & 18.94 & 0.52 & 0.81  \\
Instant3dit & 23.11 & 0.14 & 0.96\\
\rowcolor{tabours}
Ours & \textbf{26.91} & \textbf{0.09} & \textbf{0.97}  \\
\bottomrule[1pt]
\end{tabular}
\label{fan_res}
\end{minipage}
\hfill
\begin{minipage}[t]{0.48\textwidth}
\centering
\small
\setlength{\tabcolsep}{6pt}
\caption{\textbf{Generalization Ability on Physically-simulated Dataset~\cite{BBD_dataset}.}}
\begin{tabular}{l | c c c}
\toprule[2pt]
\rowcolor{tabhead}
Method & PSNR $\uparrow$ & LPIPS  $\downarrow$ & SSIM  $\uparrow$\\
\midrule[1pt]
SD & 12.02 & 0.74 & 0.53  \\
ControlNet & 14.50 & 0.59  & 0.71\\
NeRFiller & 17.66 & 0.52  & 0.79 \\
Instant3dit & 22.27 & 0.15 & \textbf{0.95} \\
\rowcolor{tabours}
Ours & \textbf{25.09} & \textbf{0.10}  & \textbf{0.95}\\
\bottomrule[1pt]
\end{tabular}
\label{qu_bbd}
\end{minipage}
\end{table*}

\begin{table*}[t]
\centering
\begin{minipage}[t]{0.48\textwidth}
\small
\centering
\setlength{\tabcolsep}{4pt}
\caption{\textbf{Ablation Studies of Multi-view Inpainting.} All variants share the base incomplete-image input (IF). Conv: concatenate noise \& incomplete image via a learnable conv layer; DMR: depth-aware mask rectifier.}
\begin{tabular}{c c | c c c}
\toprule[2pt]
\rowcolor{tabhead}
Conv & DMR & PSNR $\uparrow$ & LPIPS $\downarrow$ & SSIM $\uparrow$\\
\midrule[1pt]
\xmark & \xmark & 22.65 & 0.14 & 0.90 \\
\cmark & \xmark & 26.53 & 0.08 & 0.94 \\
\rowcolor{tabours}
\cmark & \cmark & \textbf{29.44} & \textbf{0.06} & \textbf{0.95} \\
\bottomrule[1pt]
\end{tabular}
\label{inpainting_ab}
\end{minipage}
\hfill
\begin{minipage}[t]{0.48\textwidth}
\small
\centering
\setlength{\tabcolsep}{3pt}
\caption{\textbf{Ablation Studies of Multi-view Reconstruction.} LRM: coarse-mesh initialization (unchecked\,$=$\,sphere); GR/TR: geometry/texture refinement.}
\begin{tabular}{c c c | c c c c}
\toprule[2pt]
\rowcolor{tabhead}
LRM & GR & TR & PSNR $\uparrow$ & LPIPS $\downarrow$ & CD $\downarrow$ & F-Score $\uparrow$\\
\midrule[1pt]
\cmark & \xmark & \xmark & 20.60 & 0.11 & 0.006 & 0.321 \\
\xmark & \cmark & \xmark & - & - & 0.02 & 0.197 \\
\cmark & \cmark & \xmark & - & - & \textbf{0.005} & \textbf{0.389} \\
\rowcolor{tabours}
\cmark & \cmark & \cmark & \textbf{23.35} & \textbf{0.09} & \textbf{0.005} & \textbf{0.389} \\
\bottomrule[1pt]
\end{tabular}
\label{rec_ab}
\end{minipage}
\end{table*}

\noindent\textbf{Quantitative Comparison.} 
As illustrated in Table~\ref{inpainting_res}, we observe the following: \textbf{1)} Our approach achieves the best performance in restoring shape and texture.
\textbf{2)} When applying depth images predicted by Depth-Anything~\cite{depth_anything_v2}, our method yields results comparable to those obtained with ground truth depths. 
\textbf{3)} The compared methods produce noticeably inferior results in terms of inpainting quality.
The improvement of Instant3dit when using our predicted masks (Table~\ref{inpainting_res}) indicates that accurate restoration-region perception is crucial for broken-object restoration. This observation motivates our mask self-perceiver, which removes the need for manually specified multi-view masks.

\noindent\textbf{Generalization Ability.}
1. \textit{Physically simulated broken objects.} As shown in Fig.~\ref{fig:first_f} and Table~\ref{qu_bbd}, we further test our model on the Breaking Bad Dataset~\cite{BBD_dataset}, synthesized by a physically based method that simulates the natural destruction process of geometric objects. 
2. \textit{Real-world broken objects.} As shown in Fig.~\ref{fig:first_f} and Table~\ref{fan_res}, we also evaluate our model on Fantastic Breaks~\cite{Lamb_2023_CVPR}.
These experiments demonstrate the generalization ability of our model to both \textbf{unseen} real-world scenarios and physically simulated cases, suggesting promising practical applicability, despite being trained solely on synthetic data.

\subsection{Reconstruction Results.}

\noindent\textbf{Baselines.} We compare our method against both single-view and multi-view LRMs, including LRM \cite{openlrm, hong2023lrm} and InstantMesh \cite{xu2024instantmesh}, Unique3D~\cite{wu2024unique3d}. We also compare our method with image-to-3D generation methods, Direct3D~\cite{direct3d}, Trellis~\cite{xiang2024structured} and Hunyuan3D-2\cite{zhao2025hunyuan3d20scalingdiffusion}. For single-view baselines, we input the front-view image. All of the methods use our inpainted and enhanced images as input for a fair comparison.

\noindent\textbf{Quantitative \& Qualitative Comparison.} As shown in Table~\ref{rec_res}, our method achieves superior rendered image quality and geometry accuracy, with a substantial improvement over baseline methods. In Fig.~\ref{rec_ql}, it is evident that our approach delivers clearer details and the most accurate geometry among the compared methods. To isolate the reconstruction stage, the compared reconstruction models are evaluated under the same restored-view input setting.
\noindent\textbf{Training time.} Our approach is highly efficient, requiring  20 seconds per object for geometry and texture refinements.

\subsection{Ablation Study}
\noindent\textbf{Multi-view Inpainting.} 
We conduct ablation studies on the proposed multiview Inpainting module in the following components: \textbf{1) IF.} Only inputting incomplete images into the cross-attention layers.
\begin{figure}[t]
		\centering
		\includegraphics[width=0.45\textwidth]{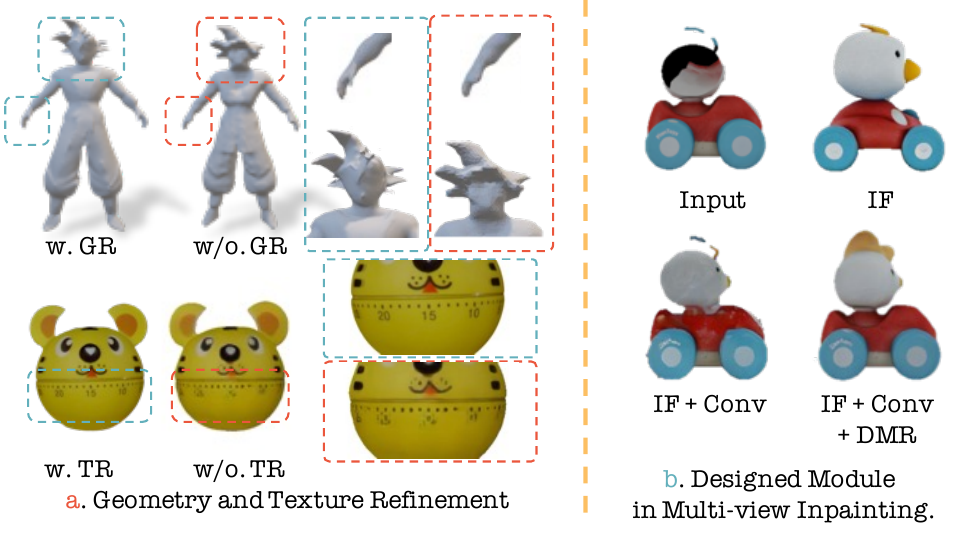}
		\caption{
            \textbf{Visualization of Ablation Studies.}
		}
		\label{rec_ab_vis}
\end{figure}
2) \textbf{Conv. } Concatenating noise and incomplete images to a learnable convolutional layer.
3) \textbf{DMR.} Adding the designed Depth-aware Mask Rectifier.
As shown in Table~\ref{inpainting_ab}, the results improve progressively with each added component, and using all designed components achieves the highest results.
As shown in Fig.~\ref{rec_ab_vis} b, 1) IF Only: the model captures the general style of the object but lacks an understanding of spatial relationships and structure.
2) IF + Conv: This enables the model to capture spatial positioning and understand object structure. However, it is still prone to color inaccuracies, especially in areas like the head (blended with the error black color). Additionally, the region that needs to be preserved is changed. 
3) IF + Conv + DMR: This allows the model to improve its ability to handle occlusions and spatial relationships, producing the best inpainting quality, with coherent colors and well-preserved spatial structure.

 \begin{figure}[!t]
		\centering
		\includegraphics[width=\linewidth]{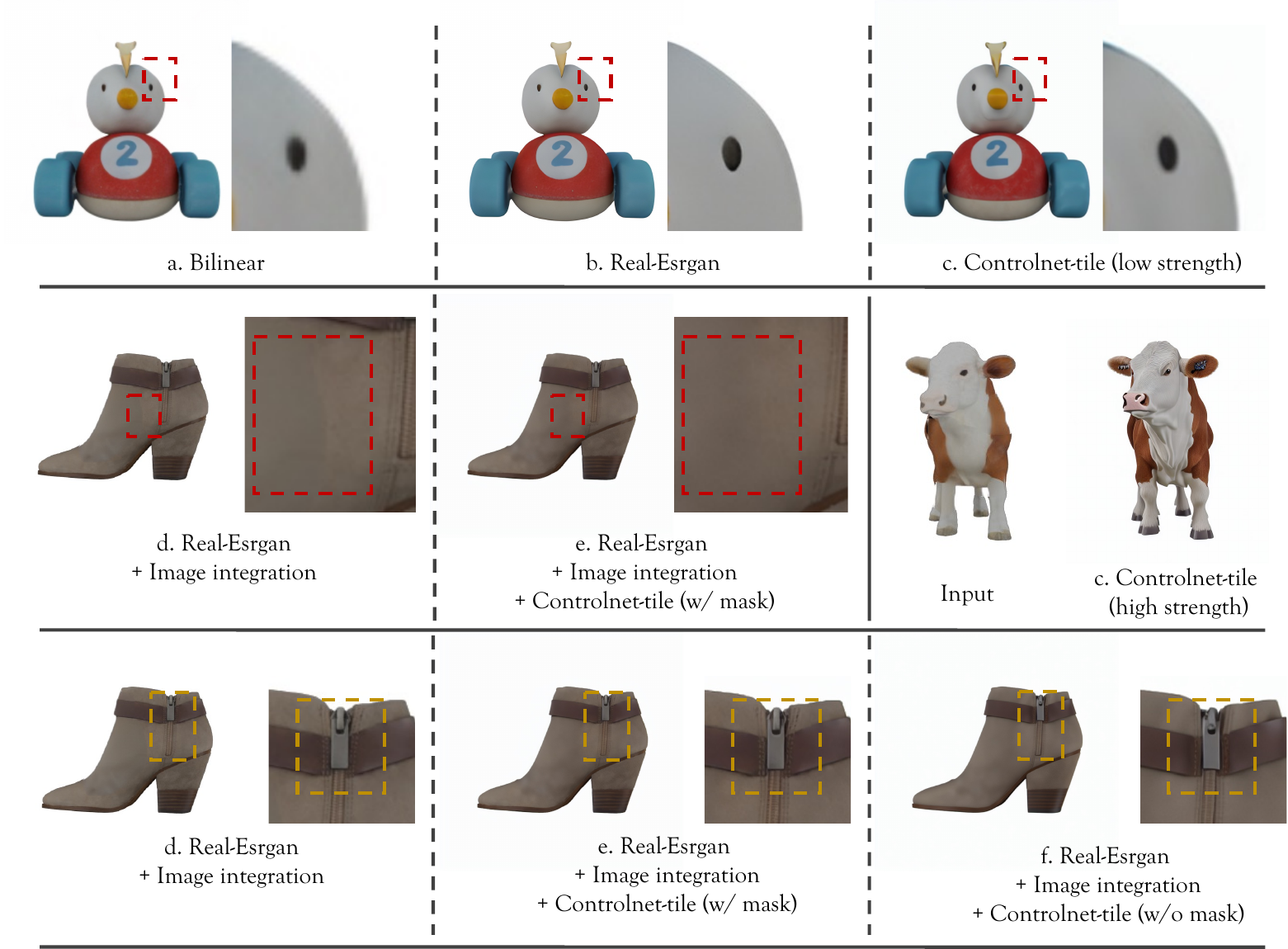}
		\caption{
            \textbf{The Effects of Image Integration and Enhancement}
            }
		
		\label{ig_en}
\end{figure}

\begin{table}[t]
\small
\centering
\setlength{\tabcolsep}{4pt}
\caption{\textbf{Ablation Studies of Image Integration and Enhancement} (256px\,$\to$\,1024px).
RE: Real-ESRGAN, II: Image Integration, CT: ControlNet-tile, MB: Mask Blending. Variant~(e) is our default.}
\begin{tabular}{c | c c c c | c c c}
\toprule[2pt]
\rowcolor{tabhead}
\# & RE & II & CT & MB & PSNR $\uparrow$ & LPIPS $\downarrow$ & SSIM $\uparrow$\\
\midrule[1pt]
a & \xmark & \xmark & \xmark & \xmark & 26.83 & 0.10 & 0.97 \\
b & \cmark & \xmark & \xmark & \xmark & 26.59 & 0.08 & 0.97 \\
c & \xmark & \xmark & \cmark & \xmark & 26.56 & 0.08 & 0.96 \\
\rowcolor{tabours}
d & \cmark & \cmark & \xmark & \xmark & \textbf{27.13} & \textbf{0.06} & \textbf{0.97} \\
e & \cmark & \cmark & \cmark & \cmark & 26.94 & \textbf{0.06} & \textbf{0.97} \\
f & \cmark & \cmark & \cmark & \xmark & 26.55 & 0.07 & \textbf{0.97} \\
\bottomrule[1pt]
\end{tabular}
\label{ig_en_tab}
\label{enhance_pipe_table}
\end{table}

\noindent\textbf{Image Integration and Enhancement.} We conduct a more detailed ablation study as shown in the Table~\ref{ig_en_tab}. 
\textbf{(a)} Baseline (Bilinear Upsampling). \textbf{(b)}. 4x Real-ESRGAN. \textbf{(c)}. 4x Controlnet-tile. \textbf{(d)} Real-ESRGAN + Image Integration. \textbf{(e)}. Real-ESRGAN + Image Integration + Controlnet-tile (w/ mask blending). \textbf{(f)}. Real-ESRGAN + Image Integration + Controlnet-tile (w/o mask blending)

As shown in Fig.~\ref{ig_en}, we observed that: 
1. Solely applying enhancement methods does not improve the quantitative metrics, but can improve visual quality. 2. The performance gains mainly originate from the image integration, which also validates that our rectified mask well indicates the regions requiring inpainting or preservation.

Overall, the organization of this stage is flexible. The key ideas are:
1. Use ControlNet-Tile with a mask-blending strategy to eliminate color inconsistencies during image integration.
2. Upsample images to the desired resolution using Real-ESRGAN, either before or after the integration.

 \noindent\textbf{Reconstruction.}
 We evaluate the impact of the following components: 1) Geometry Refinement~(GR), and 2) Texture Refinement~(TR). In Table~\ref{rec_ab} and Fig.~\ref{rec_ab_vis} a, incorporating GR leads to substantial improvements in geometry quality.  
 TR improves the visual quality of rendered images.
 
\noindent\textbf{The Effect of Coarse Meshes.}
Without LRMs, a typical alternative is to start from a simple primitive (e.g., a sphere) and optimize its shape using our geometry losses. As reported in Table~\ref{rec_ab} (LRM vs.\ sphere initialization), LRMs provide a much better initialization, leading to faster convergence and improved reconstruction quality.

 \begin{figure}[!tb]
		\centering
		\includegraphics[width=0.45\textwidth]{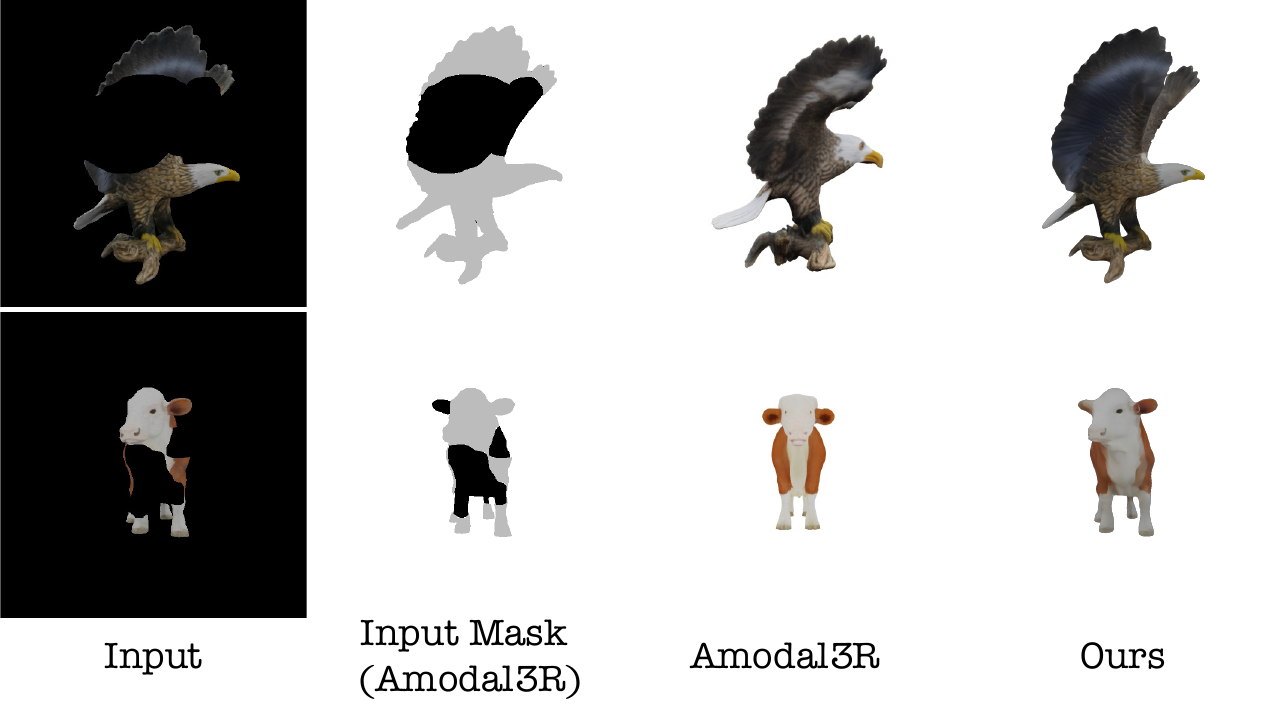}
		\caption{
            \textbf{Comparison with Amodal3R~\cite{wu2025amodal3ramodal3dreconstruction}.}}
		
		\label{cp_amodal}
\end{figure}

\begin{figure*}[!tb]
  \centering
  
    \centering
		\includegraphics[width=1.0\textwidth]{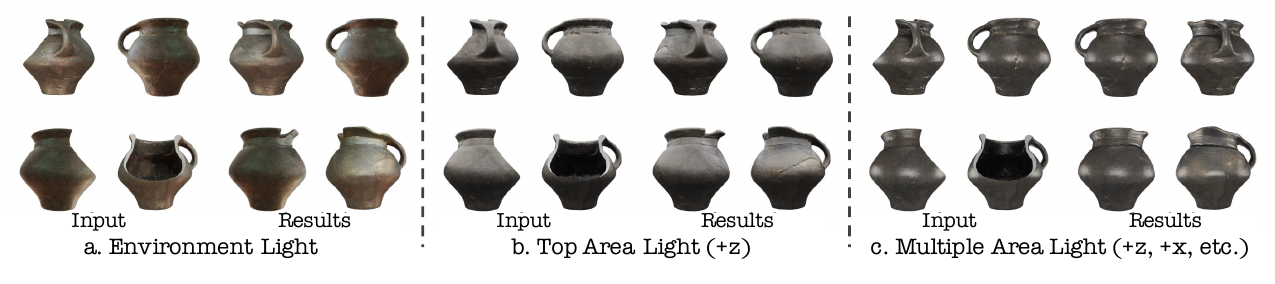}
		\caption{
            \textbf{Visualization under Different Lighting Conditions.}
		}
  
		\label{lgt_ab_vis}
\end{figure*}

\begin{table*}[!t]
\centering
\begin{minipage}[t]{0.36\textwidth}
\centering
\setlength{\tabcolsep}{4pt}
\caption{\textbf{Different Lighting Settings.}}
\begin{tabular}{l | c c c}
\toprule[2pt]
\rowcolor{tabhead}
Setting & PSNR $\uparrow$ & LPIPS $\downarrow$ & SSIM $\uparrow$\\
\midrule[1pt]
Top area light & 25.18 & 0.06 & 0.95 \\
Multiple area lights & 25.50 & 0.06 & 0.95 \\
Environment light & 25.28 & 0.06 & 0.95 \\
\bottomrule[1pt]
\end{tabular}
\label{light_ab}

\vspace{.5em}

\caption{\textbf{More Generated Views.}}
\vspace{-.5em}
\begin{tabular}{l | c c c c}
\toprule[2pt]
\rowcolor{tabhead}
Setting & PSNR $\uparrow$ & LPIPS $\downarrow$ & FID $\downarrow$ & SSIM $\uparrow$\\
\midrule[1pt]
4-view & 25.50 & 0.06 & 31.82 & 0.95 \\
6-view & 25.00 & 0.07 & 24.70 & 0.95 \\
8-view & 25.17 & 0.07 & 20.49 & 0.95 \\
\bottomrule[1pt]
\end{tabular}
\label{n_v}
\end{minipage}
\hfill
\begin{minipage}[t]{0.32\textwidth}
\centering
\small
\setlength{\tabcolsep}{6pt}
\caption{\textbf{View-consistency Scoring.}}
\begin{tabular}{l | c }
\toprule[2pt]
\rowcolor{tabhead}
Method & MEt3R \\
\midrule[1pt]
SD	& 0.44 \\
Controlnet	&0.53 \\
Nerfiller	&0.50 \\
Pix2gestalt	&0.41 \\
Instant3dit	&0.34 \\
\rowcolor{tabours}
Ours	&0.32 \\
Ground Truth	&0.29 \\
\bottomrule[1pt]
\end{tabular}
\label{vcs}
\end{minipage}
\hfill
\begin{minipage}[t]{0.30\textwidth}
\centering
\small
\setlength{\tabcolsep}{6pt}
\caption{\textbf{User Studies.}}
\begin{tabular}{l | c  c}
\toprule[2pt]
\rowcolor{tabhead}
Method & Geometry & Texture \\
\midrule[1pt]
Open-LRM	&1.9	&2.1 \\
InstantMesh	&3.2	&3.2 \\
Unique3D	&3.3	&3.5 \\
Direct3D	&3.0	&- \\
Trellis	&3.1	&3.0 \\
Hunyuan3D	&3.5	&3.6 \\
\rowcolor{tabours}
Ours	&3.9	&4.0 \\
\bottomrule[1pt]
\end{tabular}
\label{up}
\end{minipage}
\end{table*}

\subsection{Byproducts}
Our Restore3D can be directly used for some applications:
\noindent\textbf{Occluded Object Handling.} We arrange either a single object or four objects to create occluded scenarios with one view and four views, respectively, based on our 350 test samples. As shown in Fig.~\ref{fig:first_f}, the one-view occlusion scenario achieves higher performance (PSNR 27.16, LPIPS 0.06, SSIM 0.95), as the occluded regions can be inferred more easily from the visible areas. When applying four-view occlusion, our model still demonstrates strong performance (PSNR 25.62, LPIPS 0.07, SSIM 0.95).

We also compare with Amodal3R~\cite{wu2025amodal3ramodal3dreconstruction} in this setting. As shown in Figure~\ref{cp_amodal}, we find that the results of Amodal3R often misalign the conditioned images and masks.
Furthermore, we notice that the base model used by Amodal3R (Trellis) also faced similar issues with misalignment and inconsistencies, which in turn affected its ability to generate accurate completions. 


\noindent\textbf{3D Object Editing.}
We can position a cube or sphere over the target region for editing and use a Boolean operation to segment the object. This enables us to render the object as an incomplete image. We then process them using our inpainting model with a text prompt for editing. Finally, we apply the reconstruction model. In Fig.~\ref{fig:first_f}, our approach successfully handles simple editing scenarios.

  
  


\subsection{Different Lighting Settings}
We render our test samples with different lights and test our inpainting model on these rendered images.
In Table~\ref{light_ab} and Fig.~\ref{lgt_ab_vis}, the results show our model can achieve promising results under different lighting settings.

\begin{figure*}[!tb]
  \centering
    \centering
		\includegraphics[width=1.0\textwidth]{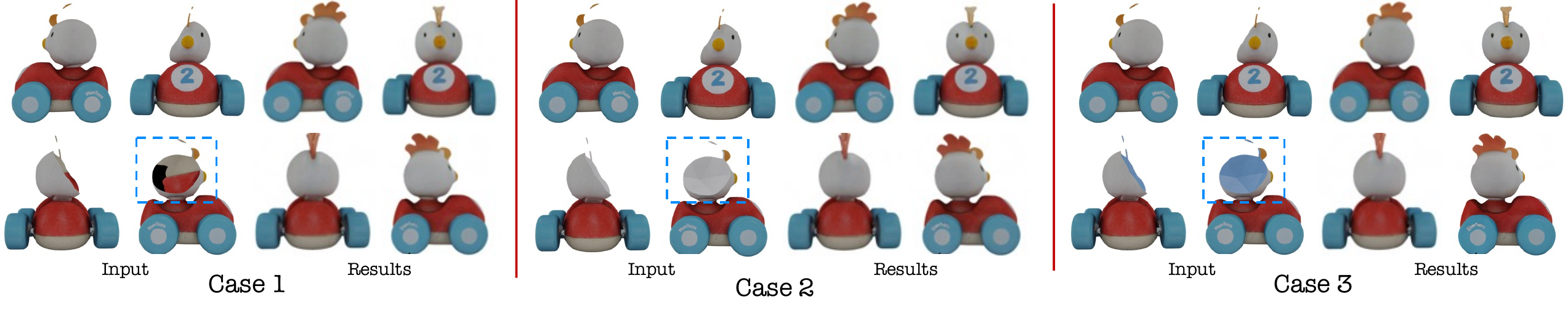}

		\caption{
            \textbf{Different Colors on Broken Planes.}
		}

		\label{diff_col}
  \end{figure*}

  \subsection{Different Colors on Broken Planes}
As shown in Fig.~\ref{diff_col}, altering the broken plane (blue dotted box) with different colors does not affect our model's ability to complete the broken objects. This further validates that our model effectively distinguishes between regions that need to be preserved and those that require generation.

\subsection{More Generated Views}
Although our model is trained on a 4-view setting, it can be directly used to process inputs with more views. As shown in Table~\ref{n_v}, the results show that their performance is comparable to the 4-view setting.

\subsection{Generated Mask Quality}

\begin{figure}[!tb]
		\centering
		\includegraphics[width=0.45\textwidth]{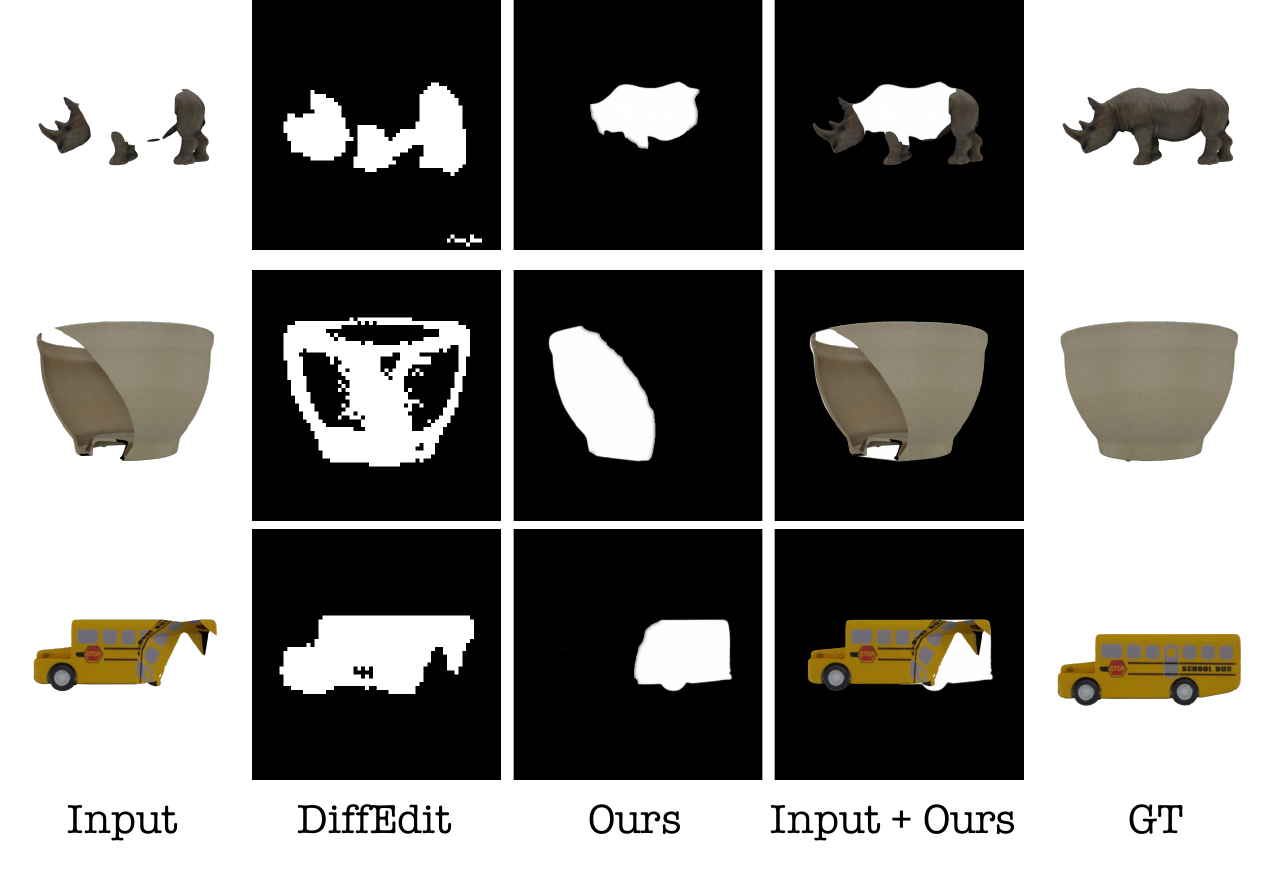}
		\caption{
            \textbf{Generated Mask Quality.}}
		
		\label{mask_gen}
\end{figure}

As shown in Figure~\ref{mask_gen}, our method significantly outperforms DiffEdit~\cite{couairon2022diffeditdiffusionbasedsemanticimage} in terms of the quality of generated masks.
DiffEdit relies solely on the difference between the noise-conditioned and unconditioned text to infer the mask. However, this approach is suboptimal because it does not account for explicit image and depth information, which are crucial for guiding the model to generate more accurate, contextually appropriate masks in the object restoration task. In contrast, our method incorporates both the image and depth as conditions, significantly improving the quality of mask generation. 

\begin{figure}[!tb]
    \centering
        \includegraphics[width=1.0\linewidth]{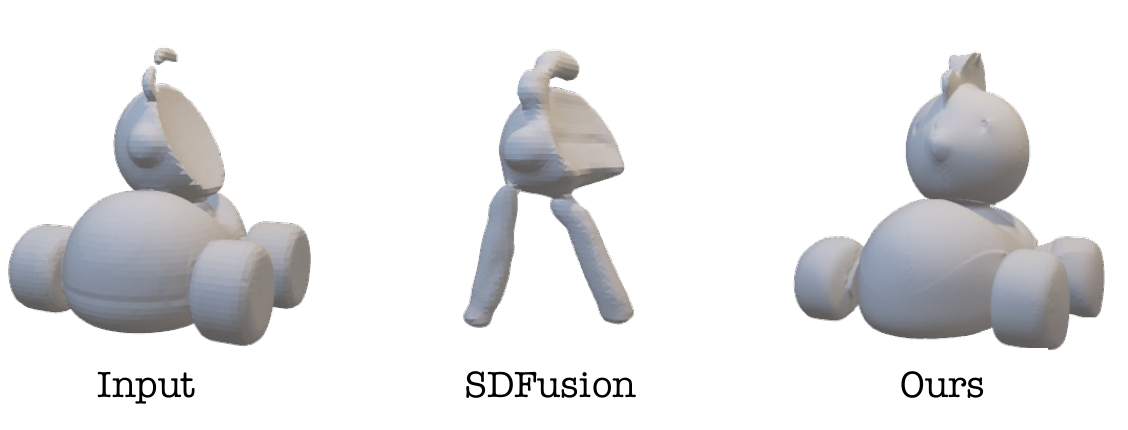}
        \caption{\textbf{Visual Comparison with SDFusion~\cite{cheng2023sdfusion}}}
        \label{vis_shape}
\end{figure}
        
\subsection{Comparison with Shape Completion Methods} We compare our method with SDFusion~\cite{cheng2023sdfusion}, a shape completion method. As shown in Fig.~\ref{vis_shape}, our method substantially outperforms SDFusion, reducing Chamfer Distance from $0.015$ to $\mathbf{0.005}$ and improving F-score from $0.096$ to $\mathbf{0.389}$.

   \begin{figure}[!tb]
      \centering
              \includegraphics[width=\linewidth]{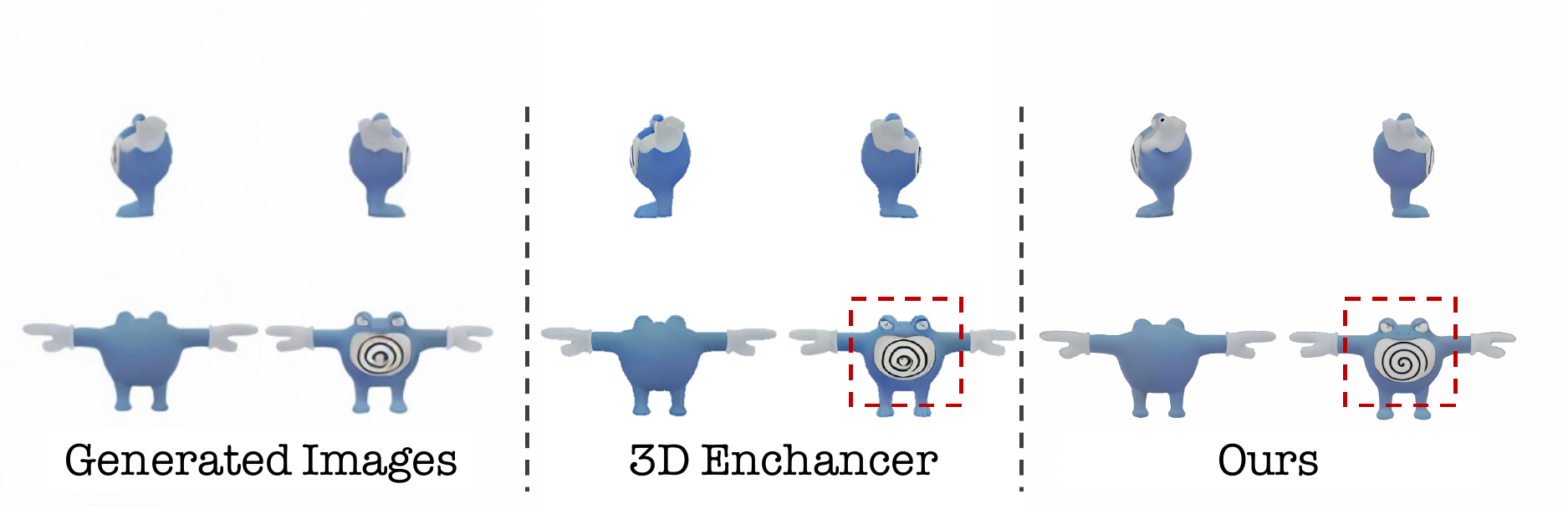}
          
              \caption{
              \textbf{Visual Comparison with 3DEnhancer~\cite{luo20253denhancerconsistentmultiviewdiffusion}}.
              }
    
      \label{enchancer}
    \end{figure}

\subsection{Image Integration and Enhancement}
As shown in Fig.~\ref{enchancer}, we provide some results of this pipeline. The results show that the proposed pipeline restores the original pattern and improves the image quality. We also compare 3DEnhancer with our image integration and enhancement pipeline, and observe superior performance: our method reaches a PSNR of $26.97$, an LPIPS of $0.05$, and an SSIM of $0.97$, compared with $26.37$, $0.07$, and $0.96$ for 3DEnhancer. It is worth noting that 3DEnhancer is limited to a resolution of 512 pixels, whereas our method supports resolutions of 1K and can be scaled up to 2K and even 4K. Moreover, 3D Enhancer can not preserve the texture of the original broken part if it does not use our predicted mask.

\begin{figure*}[!tb]
		\centering
		\includegraphics[width=0.95\textwidth]{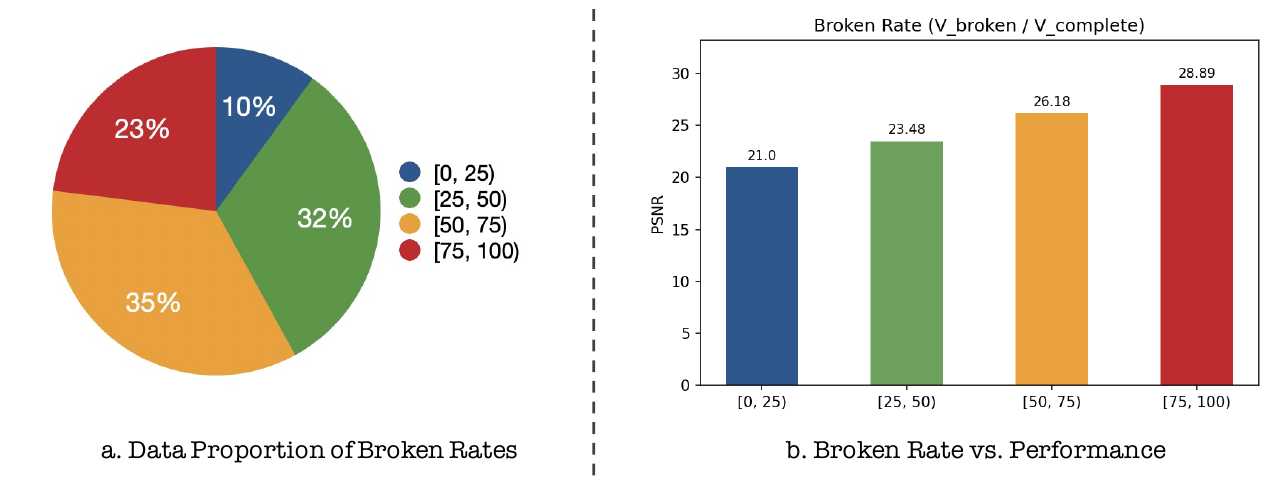}
		\caption{
            \textbf{Broken Rate vs. Performance.}}
		
		\label{fig:broken_rate}
\end{figure*}


\subsection{Fragment-size Distribution vs. Performance}
We present the fragment-size distribution and corresponding performance on our synthetic test datasets (GSO, Objaverse, OmniObject3D) and unseen datasets (Breaking Bad Dataset, Fantastic Breaks) in Fig.~\ref{fig:broken_rate}. Across all datasets, performance consistently improves as the missing region becomes smaller. This is expected, as smaller missing regions provide the model with richer contextual information, enabling more accurate inference of the missing shape.

Moreover, when the missing region becomes very large, the network gains more flexibility in generating plausible content. In such cases, the output may deviate from the original ground truth, but this discrepancy should not necessarily be considered an "error." This is because extremely large missing regions often provide little or no contextual guidance, leading to inherently ambiguous reconstructions.

\subsection{View-consistency Scoring. \& User Studies}
As shown in Table~\ref{vcs}, we use MEt3R to measure the multiview inpainted images. The table shows that our model outperforms other methods and is very close to the Ground Truth, further validating its effectiveness. We also provide user studies to measure the alignments of reconstructed meshes and original meshes, as shown in Table~\ref{up}. 5 is the best score, 1 is the worst score. The results show our model outperforms other methods.

\subsection{Compute Budgets}
Our model is computationally efficient and runs on modest GPU memory (a single NVIDIA RTX 3090 GPU with 24GB). The full pipeline takes about 44s per object, broken down as 5s for multi-view inpainting, 13s for image integration and enhancement, 6s for coarse mesh reconstruction, and 20s for geometry and texture refinement, while still delivering high-quality results.

\begin{table}[!t]
  \centering
  \caption{\textbf{Hyperparameters.}}
  \begin{tabular}{l | c}
  \toprule[2pt]
  \rowcolor{tabhead}
  Method & Value \\
  \midrule[1pt]
  Multiview inpainting Inference timestep & 50 \\
  Multiview inpainting CFG & 5.0 \\
  Controlnet-tile Inference timestep & 32 \\
  Controlnet-tile CFG & 7.5 \\
  Controlnet-tile Strength & 0.25 \\
  LRM \& StableNormal & official setting \\
  \bottomrule[1pt]
  \end{tabular}
  \label{Hyperparameters}
  \end{table}

\subsection{Hyperparameters \& Robustness}

Even though our pipeline includes a multi-stage process, our pipeline does not involve a large number of hyperparameters, making it relatively insensitive to hyperparameter choices.
Across all experiments, we do not perform any instance-specific or object-specific hyperparameter tuning; instead, we simply adopt the default or officially recommended settings. This design choice enhances both the practicality and reproducibility of our method. Furthermore, the experimental results in the ablation studies demonstrate that our model can consistently produce plausible outcomes in these standard settings.
The key hyperparameters used in our pipeline are listed in Table~\ref{Hyperparameters}.
A higher Controlnet-tile Strength leads to inconsistent or misaligned results that do not match the preserved (visible) regions. In contrast, moderate Strength values (e.g., 0.25) reliably maintain alignment while enhancing detail quality.


\section{Conclusion}
\label{conclusion}
In this paper, we propose a novel framework named Restore3D, consisting of multi-view image inpainting and reconstruction, to simultaneously complete both the shape and texture of broken 3D objects. 
To facilitate this task, we develop an automated data processing pipeline that collects pair-wise data from a large-scale dataset~\cite{deitke2023objaverse}. In the multi-view image inpainting, we design a mask self-perceiver with a depth-aware mask rectifier. This component autonomously identifies and reconstructs missing regions while preserving the original patterns. To address the low resolution resulting from the base model~\cite{shi2023mvdream}, 
we implement an image integration and enhancement pipeline, allowing for seamless integration and detail enhancement by learned masks. 
For the reconstruction stage, we employ an LRM to quickly generate a coarse result, followed by separate geometry refinement using normal priors and texture refinement using enhanced images. 
Through this designed framework, our model produces coherent completions of broken objects as illustrated in Fig.~\ref{fig:first_f}. Moreover, our designed framework can also handle simple 3D object editing and occluded objects.

\noindent\textbf{Limitations.}
\label{limi}
Our approach builds upon a base model and thus inevitably inherits some of its limitations. For instance, the low resolution of the input restricts the ability to capture very fine details, such as the facial features of characters, even with the application of enhancement techniques. In addition, there is still a lot of room to enrich the quality of geometry and material details in the reconstruction.


%



\ifCLASSOPTIONcaptionsoff
  \newpage
\fi



\small{
\bibliographystyle{IEEEtran}
\bibliography{mybib}
}
%


\end{document}